\documentclass[10pt,twocolumn,letterpaper]{article}

\usepackage{cvpr}              

\usepackage{graphicx}
\usepackage{amsmath}
\usepackage{amssymb}
\usepackage{booktabs}
\usepackage{bbm}

\usepackage[pagebackref,breaklinks,colorlinks]{hyperref}

\usepackage[capitalize]{cleveref}
\crefname{section}{Sec.}{Secs.}
\Crefname{section}{Section}{Sections}
\Crefname{table}{Table}{Tables}
\crefname{table}{Tab.}{Tabs.}

\def\cvprPaperID{*****} 
\def\confName{OOD-CV}
\def\confYear{2022}

\begin{document}

\title{Feed-Forward Source-Free Domain Adaptation via Class Prototypes}

\author{Ondrej Bohdal\\
The University of Edinburgh \thanks{Work from an internship at Samsung AI Center, Cambridge, UK.}\\
Edinburgh, UK\\
{\tt\small ondrej.bohdal@ed.ac.uk}
\and
Da Li\\
Samsung AI Center\\
Cambridge, UK\\
{\tt\small da.li1@samsung.com}
\and
Timothy Hospedales\\
The University of Edinburgh and\\
Samsung AI Center, UK\\
{\tt\small t.hospedales@ed.ac.uk}
}
\maketitle

\begin{abstract}
Source-free domain adaptation has become popular because of its practical usefulness and no need to access source data. However, the adaptation process still takes a considerable amount of time and is predominantly based on optimization that relies on back-propagation. In this work we present a simple feed-forward approach that challenges the need for back-propagation based adaptation. Our approach is based on computing prototypes of classes under the domain shift using a pre-trained model. It achieves strong improvements in accuracy compared to the pre-trained model and requires only a small fraction of time of existing domain adaptation methods.
\end{abstract}

\section{Introduction}
In real-world applications of machine learning it is common to encounter domain shift i.e. target deployment data come from a different distribution than source data. For example, there is a significant difference in style between images taken during the summer and winter. Domain shift typically results in decreased performance of models trained only on the source data, yet with suitable techniques it is possible to recover a significant part of the lost performance. The practical importance of tackling domain shift has led to domain adaptation becoming a popular field with a wide variety of methods and scenarios.

The standard domain adaptation scenario is that a model is trained with labelled source data alongside unlabelled target domain data. However, it has been shown recently \cite{Liang2020DoAdaptation} that access to the source data is not needed and strong performance can be achieved even with a pre-trained model and unlabelled target data only. Model trained on the source data is adapted using only unlabelled target data. This scenario has become known as \textit{source-free} domain adaptation (SFDA) and has attracted a lot of interest due to its practical usefulness. For example, storing a large amount of source domain data on an edge device such as smartphone is not practical and SFDA eliminates such need.

Existing methods for source-free domain adaptation predominantly use back-propagation based fine-tuning that takes many iterations. In our work we challenge the need for these costly methods and instead focus on feed-forward methods. A simple example of feed-forward SFDA is to update the batch normalisation statistics \cite{Ishii2021Source-freeStatistics}. We propose an alternative simple feed-forward strategy that achieves significantly better results.

Inspired by prototypical networks from few-shot learning literature \cite{Snell2017PrototypicalLearning}, we propose to construct prototypes of different classes under the domain shift, using the pre-trained model for extracting features and pseudo-labels. During inference we find the closest prototype to the features of the current test example, and we select the class of that prototype as prediction. We replace the classifier layer by a nearest-prototype layer, which can be understood as a new way of fine-tuning pre-trained models.

Our results on a variety of standard benchmarks -- Office, Office-Home and ViSDA-C -- show this is a strategy that can lead to surprisingly strong results, at a very small fraction of time compared to standard SFDA methods. We argue our method is a simple baseline which should be included in the comparison when evaluating various SFDA methods. At the same time it is a practical strategy that can be applied on edge devices with great success. Our main contribution is a new simple and fast feed-forward method for source-free domain adaptation.

\section{Related Work}
\textbf{Source-free domain adaptation: } After \cite{Liang2020DoAdaptation} observed unsupervised domain adaptation does not need access to the source data, the field of SFDA has attracted a lot of attention and several methods were proposed. \cite{Liang2020DoAdaptation} have developed two methods to solve the practical scenario they have identified, namely SHOT-IM and SHOT methods. SHOT-IM is a simpler version of SHOT, and it fine-tunes the pre-trained feature extractor by maximizing information transferred from source model. SHOT extends SHOT-IM by self-supervised pseudo-labelling, which leads to improved results but makes the method more costly. Other recent methods include universal SFDA (USFDA) \cite{Kundu2020UniversalAdaptation} that performs SFDA using an instance-level weighting mechanism called source similarity metric, and historical contrastive learning (HCL) \cite{Huang2021ModelData} that uses historical contrastive instance and category discrimination. SFDA can also be used for semantic segmentation \cite{Liu2021Source-freeSegmentation, Kundu2021GeneralizeSegmentation} and object detection \cite{Li2020ModelData}.

\textbf{Test-time domain adaptation: } Related to SFDA is test-time domain adaptation (TTDA) that does adaptation on the current test minibatch, rather than using a separate dataset for adaptation. From these, TENT \cite{Wang2021Tent:Minimization} optimizes channel-wise affine transformation using the current minibatch, while ARM \cite{Zhang2021AdaptiveShift} provides a general meta-learning framework for TTDA that includes for example updates to batch normalisation statistics or context network that embeds the whole minibatch and uses it as additional channels. Test-time template adjuster (T3A) \cite{Iwasawa2021Test-timeGeneralization} is another recent method for TTDA and also uses the idea of prototypes in combination with entropy-based filtering. However, \cite{Iwasawa2021Test-timeGeneralization} have not studied the use of prototypes for source-free domain adaptation and their approach towards constructing prototypes is different from ours. They consider a streaming scenario where examples arrive and are classified with prototypes constructed with examples seen so far.

\section{Methods}
In the SFDA scenario, we are given a pre-trained model, composed of feature extractor $f_{\theta}$ and classifier $f_{\phi}$, and target dataset $\mathcal{D}_T$. The goal is to adapt the pre-trained model and obtain strong performance on $\mathcal{D}_T$. As part of our method we construct a prototype of each class and use the prototypes for classification instead of the pre-trained classifier layer.

We construct the prototypes $\boldsymbol{c}_l$ of different classes $l$ as 
\begin{equation} \label{eq:proto}
\boldsymbol{c}_l=\frac{\sum_{i=1}^{N} f_\theta(x_i) w_i}{\sum_{i=1}^{N} w_i},
\end{equation}

where $N=|\mathcal{D}_T|$ is the number of examples in the target dataset, and the weights $w_i$ with which the features extracted for example $x_i$ contribute to the prototype computation are: 

$$w_i = [f_{\phi \circ \theta}(x_i)]_l \mathbbm{1}(\text{argmax}f_{\phi \circ \theta}(x_i) == l).$$

Only examples predicted to be of class $l$ contribute to calculation of class prototype $\boldsymbol{c}_l$, for which we use an indicator function denoted by $ \mathbbm{1}$. The examples for which confidence is larger contribute more -- more specifically $f_{\phi \circ \theta}(x_i)$ gives probabilities of different classes and $[f_{\phi \circ \theta}(x_i)]_l$ gives us the probability of class $l$. We have found confidence-weighting to consistently improve the quality of prototypes, compared to standard one-hot approach used in prototypical networks \cite{Snell2017PrototypicalLearning} or T3A \cite{Iwasawa2021Test-timeGeneralization}. Predictions are made by extracting features from the test example and finding the closest prototype, using the selected distance measure. We name our method PDA (prototypical domain adaptation).

\textbf{PDA w MCD: } We extend PDA with an initial step that uses robust generative classifier (RoG) based on Minimum Covariance Determinant \cite{Lee2019RobustLabels} to obtain higher-quality pseudo-labels. Higher-quality pseudo-labels enable us to construct more accurate prototypes of different classes under the domain shift. RoG replaces the pre-trained classifier and is only used for creating prototypes.

In our case the parameters of RoG are estimated using the features and pseudo-labels given by the pre-trained model when applied on target domain data. We follow the approach from \cite{Lee2019RobustLabels} with the approximate version inspired by \cite{Hubert2004FastAnalysis}. We use one RoG classifier rather than an ensemble of generate classifiers because an ensemble led to only marginal improvements.

\section{Experiments}
\subsection{Experimental details}
\textbf{Benchmarks: } We use standard benchmarks used for evaluating SFDA methods, namely Office \cite{Saenko2010AdaptingDomains}, Office-Home \cite{Venkateswara2017DeepAdaptation} and VisDA-C \cite{Peng2018VisDA:Adaptation}.

\textbf{Baselines: } We compare our approach with 1) directly using the pre-trained model on the target dataset, and 2) updating the BN statistics using the target dataset. We also provide results for SHOT-IM and SHOT methods from \cite{Liang2020DoAdaptation} to give context about performance of current back-propagation based methods for SFDA. The focus is on illustrating the performance of back-propagation based methods rather than doing a comparison with all recent back-propagation based SFDA approaches.

\textbf{Experiment set-up: } We follow the set-up from \cite{Liang2020DoAdaptation} and use the code they have released. We also utilize their pre-trained models, which are provided for three random seeds. For Office and Office-Home benchmarks we use ResNet50, and for ViSDA-C we use ResNet101 \cite{He2015DeepRecognition}. We select cosine distance to find which class prototype is the closest to the current test example. Following \cite{Liang2020DoAdaptation} we report the accuracies on the target set, reporting the average across three random seeds. As part of PDA, we also update the BN statistics.

\begin{table*}[t]
\vskip 0.15in
\begin{center}
\begin{small}
\begin{sc}
\begin{tabular}{lccccccc}
\toprule
Method & A$\rightarrow$D & A$\rightarrow$W & D$\rightarrow$A & D$\rightarrow$W & W$\rightarrow$A & W$\rightarrow$D & Avg. \\
\midrule
Source & 80.8 & 77.1 & 60.2 & 95.3 & 63.5 & 98.7 & 79.3 \\
BN update & 80.1 & 80.7 & 61.2 & 97.4 & 61.9 & 99.7 & 80.2 \\
\midrule
PDA & 88.3 & 87.8 & 67.6 & 98.1 & 68.1 & 99.7 & 84.9 \\
PDA w MCD & 88.2 & 88.9 & 69.3 & 97.7 & 69.3 & 99.7 & 85.5 \\
\midrule
SHOT-IM (paper) & 90.6 & 91.2 & 72.5 & 98.3 & 71.4 & 99.9 & 87.3 \\
SHOT (paper) & 94.0 & 90.1 & 74.7 & 98.4 & 74.3 & 99.9 & 88.6 \\
\bottomrule
\end{tabular}
\caption{Office dataset - target domain accuracies (\%) across various source and target domains -- Amazon, DSLR and Webcam.}
\label{tab:office-main}
\end{sc}
\end{small}
\end{center}
\vskip -0.1in
\end{table*}

\begin{table*}[t]
\vskip 0.15in
\begin{center}
\begin{small}
\begin{sc}
\resizebox{\textwidth}{!}{\begin{tabular}{lccccccccccccc}
\toprule
Method & A$\rightarrow$C & A$\rightarrow$P & A$\rightarrow$R & C$\rightarrow$A & C$\rightarrow$P & C$\rightarrow$R & P$\rightarrow$A & P$\rightarrow$C & P$\rightarrow$R & R$\rightarrow$A & R$\rightarrow$C & R$\rightarrow$P & Avg. \\
\midrule
Source & 44.6 & 67.4 & 74.7 & 52.7 & 62.7 & 64.8 & 53.1 & 40.7 & 73.2 & 65.3 & 45.4 & 78.1 & 60.2 \\
BN update & 43.8 & 63.4 & 73.8 & 56.1 & 63.4 & 66.8 & 54.5 & 42.7 & 74.0 & 66.4 & 47.6 & 77.3 & 60.8 \\
\midrule
PDA & 48.1 & 71.2 & 77.7 & 61.9 & 71.2 & 73.4 & 60.4 & 46.2 & 78.3 & 68.8 & 50.4 & 80.2 & 65.6 \\
PDA w MCD & 49.1 & 72.4 & 78.6 & 63.0 & 72.9 & 74.5 & 61.0 & 47.9 & 79.5 & 69.5 & 51.5 & 80.6 & 66.7 \\
\midrule
SHOT-IM (paper) & 55.4 & 76.6 & 80.4 & 66.9 & 74.3 & 75.4 & 65.6 & 54.8 & 80.7 & 73.7 & 58.4 & 83.4 & 70.5 \\
SHOT (paper) & 57.1 & 78.1 & 81.5 & 68.0 & 78.2 & 78.1 & 67.4 & 54.9 & 82.2 & 73.3 & 58.8 & 84.3 & 71.8 \\
\bottomrule
\end{tabular}}
\caption{Office-Home dataset - target domain accuracies (\%) across various source and target domains -- Art, Clipart, Product, Real-World.}
\label{tab:office-home-main}
\end{sc}
\end{small}
\end{center}
\vskip -0.1in
\end{table*}

\subsection{Results}
We give our main results in Table \ref{tab:office-main} for Office benchmark, Table \ref{tab:office-home-main} for Office-Home benchmark and Table \ref{tab:visda-c-main} for ViSDA-C synthetic-to-real benchmark. We observe that for Office and Office-Home simply updating BN statistics is not sufficient, but with PDA we achieve a significant increase in accuracy compared to the pre-trained source model. As a result, we are able to get closer to the performance of standard back-propagation based methods, at a very small fraction of time. Combination of PDA with MCD leads to additional improvements in accuracy of PDA. We also observe significant accuracy improvements on large-scale ViSDA-C benchmark.

\begin{table}[h!]
\vskip 0.15in
\begin{center}
\begin{small}
\begin{sc}
\begin{tabular}{lc}
\toprule
Method & Syn-to-Real \\
\midrule
Source & 46.6 \\
BN update & 64.7 \\
\midrule
PDA & 70.9 \\
PDA w MCD & 72.0 \\
\midrule
SHOT-IM (paper) & 80.4 \\
SHOT (paper) & 82.9 \\
\bottomrule
\end{tabular}
\caption{SFDA on ViSDA-C synthetic-to-real - target domain accuracies (\%). Using official pre-trained models and averaged across three seeds. ResNet101.}
\label{tab:visda-c-main}
\end{sc}
\end{small}
\end{center}
\vskip -0.1in
\end{table}

\section{Analysis}
As part of analysis we study several questions: 1) How does the adaptation time of PDA (or PDA w MCD) compare with standard back-propagation based methods? 2) What is the estimated upper bound performance when using true labels for constructing prototypes? 3) Would applying RoG classifier trained under domain shift be sufficient?

\begin{figure}[h!]
\vskip 0.2in
\begin{center}
\centerline{\includegraphics[width=\columnwidth]{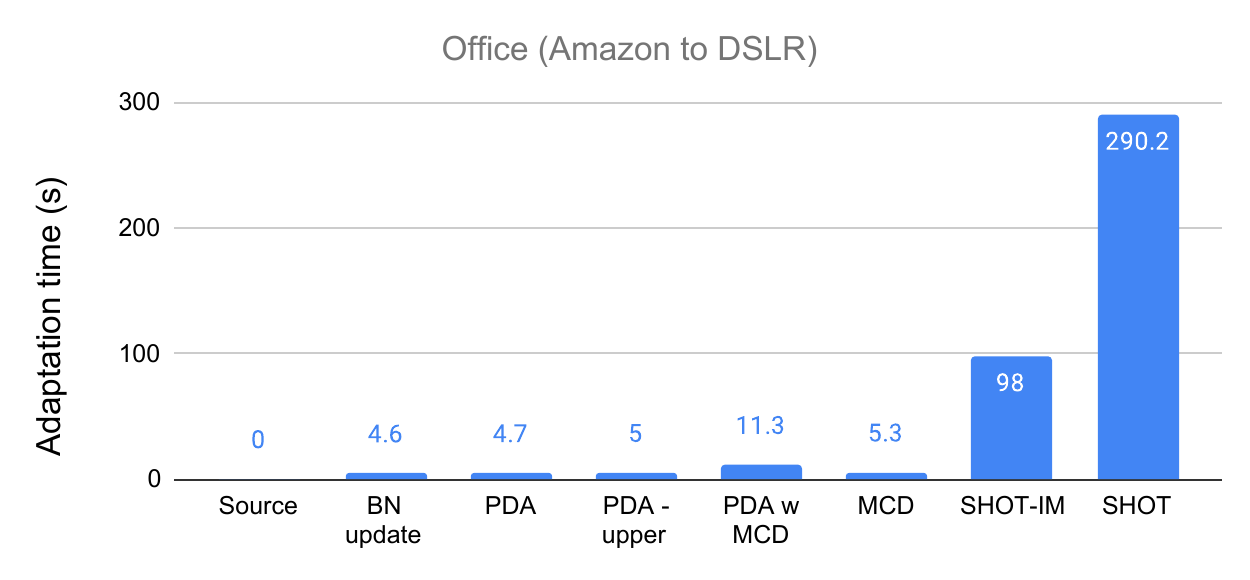}}
\centerline{\includegraphics[width=\columnwidth]{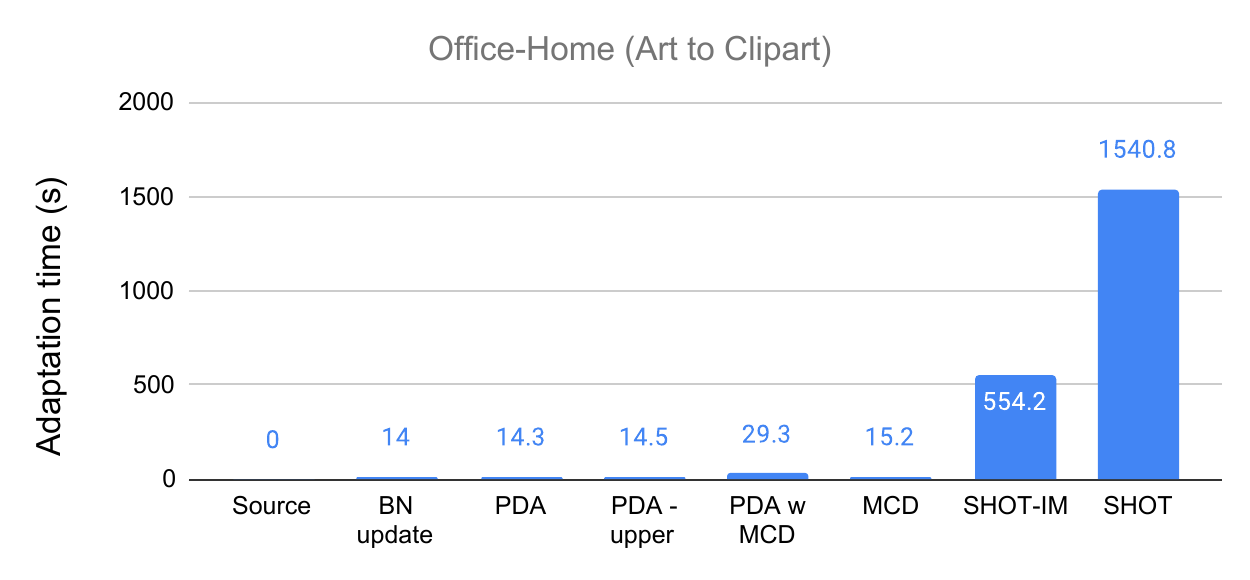}}
\caption{Adaptation time on Office and Office-Home datasets for selected source and target domains.}
\label{fig:time}
\end{center}
\vskip -0.2in
\end{figure}

\textbf{Time comparison: } We compare our PDA method in terms of adaptation time with existing back-propagation based methods to show the clear speed benefits our methods bring. The results in Figure \ref{fig:time} confirm the adaptation time required by our PDA (or PDA w MCD) method is only a very small fraction of adaptation time required by SHOT-IM and SHOT methods that use back-propagation. Evaluation times (inference) have been similar for all methods.

\textbf{Upper bound of PDA: } We use the true labels of target domain data to estimate an upper bound on the performance of PDA. Results in Table \ref{tab:upper-bound} show that using true labels leads to large improvements, which suggests there is potential for development of new highly-accurate feed-forward SFDA methods. It is particularly interesting to observe that for Office and Office-Home dataset the upper bound estimated using the true labels is a few percentages higher than what is obtained by SHOT-IM and SHOT methods.

\begin{table}[h!]
\vskip 0.15in
\begin{center}
\begin{small}
\begin{sc}
\begin{tabular}{lccc}
\toprule
Method & Office & Office-Home & ViSDA-C \\
\midrule
Source & 79.3 & 60.2 & 46.6 \\
BN update & 80.2 & 60.8 & 64.7 \\
\midrule
PDA & 84.9 & 65.6 & 70.9 \\
PDA w MCD & 85.5 & 66.7 & 72.0 \\
PDA -- upper & 91.0 & 75.9 & 76.1 \\
\bottomrule
\end{tabular}
\end{sc}
\end{small}
\caption{Upper bound performance of PDA by using true labels -- average accuracy across domains (\%).}
\label{tab:upper-bound}
\end{center}
\vskip -0.1in
\end{table}

\textbf{Directly using RoG classifier based on MCD: } We have evaluated the novel use of RoG MCD-based classifier for SFDA in Table \ref{tab:mcd-direct}. The results show that even if this approach is helpful, it needs to be combined with PDA for the best results, particularly for Office dataset.

\begin{table}[h!]
\vskip 0.15in
\begin{center}
\begin{small}
\begin{sc}
\begin{tabular}{lccc}
\toprule
Method & Office & Office-Home & ViSDA-C \\
\midrule
Source & 79.3 & 60.2 & 46.6 \\
BN update & 80.2 & 60.8 & 64.7 \\
\midrule
PDA w MCD & 85.5 & 66.7 & 72.0 \\
MCD & 81.6 & 65.4 & 70.4 \\
\bottomrule
\end{tabular}
\end{sc}
\end{small}
\caption{Evaluation of using RoG classifier based on MCD directly -- average accuracy across domains (\%).}
\label{tab:mcd-direct}
\end{center}
\vskip -0.1in
\end{table}

\section{Conclusion}
We have developed a new simple method for feed-forward source-free domain-adaptation that is based on computing prototypes of different classes under domain shift. The method achieves strong results and takes only a small fraction of time to run compared to standard back-propagation based methods. Our evaluation with true labels has shown there is scope for developing further new feed-forward SFDA methods, for which our method acts as a well-performing baseline.

\newpage
{\small
\bibliographystyle{ieee_fullname}
\bibliography{references}
}

\end{document}